\documentclass[conference]{IEEEtran}
\IEEEoverridecommandlockouts
\usepackage{textcomp}
\usepackage{xcolor}
\pdfoutput=1
\usepackage{graphicx} 
\usepackage{algorithm}
\usepackage{algorithmic}

\usepackage{float} 
\usepackage{amsmath,amssymb,amsthm}
\usepackage{marvosym}
\usepackage{cite}
 
\usepackage{stfloats}
\usepackage{color}

\usepackage[english]{babel} 
\usepackage{array} 

\usepackage{makecell}

\newtheorem{theorem}{Theorem} 
 
\usepackage{siunitx}

\begin{document}

\title{Heterogeneous
Multi-agent Collaboration in UAV-assisted Mobile Crowdsensing Networks
} 
\author{
Xianyang Deng\textsuperscript{*}, Wenshuai Liu\textsuperscript{\dag}, Yaru Fu\textsuperscript{\Letter \ddag}, and Qi Zhu\textsuperscript{*}\\
    \textsuperscript{*}\textit{The Key Wireless Laboratory, Nanjing University of Posts and Telecommunications, China} \\
    \textsuperscript{\dag}\textit{The School of Artificial Intelligence and Computer Science, Jiangnan University, China} \\
    \textsuperscript{\ddag}\textit{The School of Science and Technology, Hong Kong Metropolitan University, Hong Kong} \\
    E-mails: 2022010106@njupt.edu.cn, liuws1996@gmail.com, yfu@hkmu.edu.hk, zhuqi@njupt.edu.cn 
\vspace{-0.5cm}
 	\thanks{ This work was supported in part by Jiangsu Provincial Key Research and Development Program under Grant BE2022068-2,  in part by the Team-based Research Fund under Reference No. TBRF/2024/1.10, and in part by the Postgraduate Research \& Practice Innovation Program of Jiangsu Province under Grant KYCX24\_1176.   
    This work was completed while Xianyang Deng was affiliated with the Hong Kong Metropolitan
    University. (\textit{Corresponding author: Yaru Fu})  
        } 
}
\maketitle

\begin{abstract}
Unmanned aerial vehicles (UAVs)-assisted mobile crowdsensing (MCS) has emerged as a promising paradigm for data collection.
However, challenges such as spectrum scarcity, device heterogeneity, and user mobility hinder efficient coordination of sensing, communication, and computation. 
To tackle these issues, we propose a joint optimization framework that integrates time slot partition for sensing,
communication, and computation phases, resource allocation, and UAV 3D trajectory planning, aiming to maximize the amount of processed sensing data.
The problem is formulated as a non-convex stochastic
optimization and further modeled as a partially observable
Markov decision process (POMDP) that can be solved by multi-agent deep reinforcement learning (MADRL) algorithm. To overcome the limitations of conventional multi-layer perceptron (MLP) networks, we design a novel MADRL algorithm with hybrid actor network.
The newly developed method is based on heterogeneous agent
proximal policy optimization (HAPPO), empowered by convolutional neural networks (CNN) for feature extraction and Kolmogorov-Arnold networks (KAN) to  capture structured state-action dependencies. 
Extensive numerical results demonstrate that our proposed  method achieves significant improvements in the amount of
processed sensing data when compared with other benchmarks.
\end{abstract}


\section{Introduction}
The rapid development of the internet of things (IoT) has enabled the large-scale deployment of sensors. These sensors are applied in various domains, such as intelligent transportation, environmental monitoring, smart cities, and marine observation \cite{9779853,10742568,10807189}.
However, traditional sensing systems with static sensor deployments often incur high deployment costs and provide limited flexibility in dynamic or large-scale environments.
Mobile crowdsensing (MCS) has emerged as a scalable solution that utilizes mobile users to collect and transmit sensing data, thereby enabling diverse IoT services.
Although MCS holds great promise, its performance is often constrained by reliance on terrestrial communication infrastructure, which performs poorly in remote or disaster-affected environments.
To overcome these challenges, unmanned aerial vehicles (UAVs) are increasingly adopted in MCS networks due to their high mobility and flexible deployment.
Consequently, UAV-assisted MCS provides a promising framework for reliable and efficient data collection in dynamic and infrastructure-limited environments. 
\begin{figure}[!t]
    \centering
    \includegraphics[width=7cm,height=4.4cm]{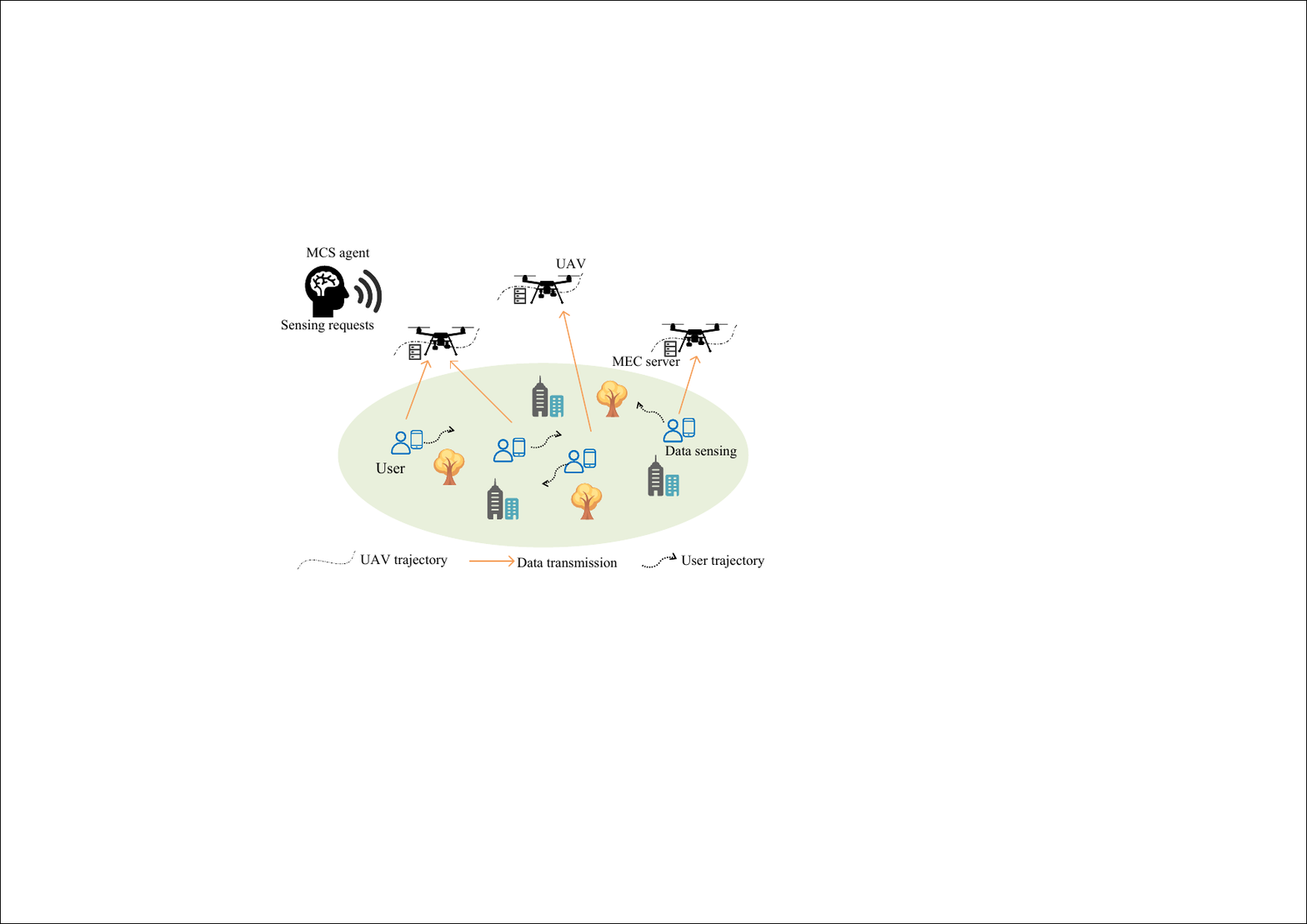}
    \caption{The multi-UAV-assisted MCS network.}
    \label{fig:example}
\end{figure}

Most recently, there appear some significant studies that focused on incorporating UAVs into MCS networks.
For instance, \cite{10731856} proposed an incentive mechanism for UAV-assisted MCS that integrates federated learning with prospect theory. 
This approach enables UAVs to securely share high-quality local models through game-theoretic optimization. 
Nevertheless, in practical deployments, the mobility of UAVs and users results in dynamic topologies and time-varying channel conditions, posing challenges for stable connectivity and reliable data collection. 
The dynamic environment and the interdependence of sensing, communication, and computation make system optimization highly challenging.
To address this, a joint sensing, communication, and computation (JSCC) framework is adopted.
In this context, \cite{9583941} proposed a cooperative data sensing and computation offloading scheme for UAV-assisted MCS networks, leveraging the computation resources of UAVs. 
By employing a deep reinforcement learning algorithm, the framework jointly optimized UAV flight trajectories, task offloading decisions, and energy consumption to maximize overall system utility.
\cite{10571223} further explored the utilization of user-side computation resources and developed a real-time optimization framework for mobile user selection, resource allocation, and UAV trajectory planning. To tackle the inherent non-convexity of the problem, the authors employed a proximal policy optimization algorithm to derive near-optimal solutions.

Although existing studies have laid a solid foundation, several critical challenges remain unsolved. 
Most  JSCC  frameworks overlook the intricate coupling between user and UAV computation, hindering efficient data aggregation and processing. 
Moreover, the presence of device heterogeneity and environmental dynamics significantly increases the complexity of time slot partition across sensing, communication, and computation phases. 
These issues are further exacerbated by user mobility and the inability of traditional optimization approaches to adapt to rapidly changing system states.

To address the aforementioned challenge, we propose a novel JSCC framework that jointly optimizes user-UAV association, resource allocation, and UAV trajectory. 
The framework fully leverages the computation resources of both users and UAVs and dynamically adjusts time slot partition to maximize the amount of processed sensing data. 
Given that the original problem is a non-convex stochastic optimization, it is reformulated as a  partially observable Markov decision process (POMDP). To solve this problem efficiently, we develop a new multi-agent deep reinforcement learning (MADRL) algorithm called CKAN-HAPPO.
The proposed method integrates convolutional neural networks (CNN) for feature extraction and a Kolmogorov-Arnold network (KAN) that utilizes spline-based adaptive activations.
This design captures complex nonlinear state-action relationships more effectively while reducing parameter complexity in high-dimensional continuous action spaces.
Furthermore, we improve training efficiency by deriving closed-form expressions that reduce the action space dimensionality of user-side computation resources.
Simulation results demonstrate that our proposed algorithm consistently outperforms existing baselines.

\section{System Model}
We consider an MCS network assisted by multiple UAVs, as depicted in Fig. 1.
The network comprises a set of users $\mathcal{K} = \left \{1,2,..., K  \right \}$ and a set of UAVs $\mathcal{U} = \left \{1,2,..., U  \right \}$.
Each user serves as a sensing service provider with data collection capabilities, while an internet-based sensing platform dispatches sensing tasks and leverages UAVs to assist in data collection.
In the model,  a time period \( T \) is divided into \( N \) equal time slots of duration \( \delta \), where \( \mathcal{N} = \{1,2,\dots,N\} \) denotes the set of time slots. 
In each time slot \( n \), a sensing task is issued by the platform. 
The slot duration is assumed to be equal to the channel coherence time, such that the relative positions between UAVs and users remain unchanged within each slot \cite{8370831}.
First, the platform schedules a group of users to participate in the sensing task. 
Each selected user then divides the sensing data into two portions: one is processed locally, while the other is offloaded to the associated UAV for computation.
To manage task execution within a slot, we define  time slot partition variables \( \xi_{k,1}[n] \) and \( \xi_{k,2}[n] \).
Specifically, user $k$ allocates $\xi_{k,1}[n]\delta$ for data sensing, while the remaining time $(1 - \xi_{k,1}[n])\delta$ is further divided into communication and computation phases. 
Within this portion, $(1 - \xi_{k,1}[n])\xi_{k,2}[n]\delta$ is used to transmit data to UAVs, and $(1 - \xi_{k,1}[n])(1 - \xi_{k,2}[n])\delta$ is assigned to UAV-side computation.
In parallel, user $k$ also performs local computation on part of the sensing data.

\subsection{Mobility Model}
\subsubsection{User Mobility Model}
The users are initially distributed randomly and move according to the Gauss-Markov model~\cite{10584314}, with the position of user $k$ in slot $n$ denoted as ${\bf q}_k[n] = (x_k[n], y_k[n], 0)$.
The movement speed and direction of user $k$ are represented as follows:
\begin{align}
&v_{k}\left[n\right]= c_{1}v_{k}\left[n-1\right] + \left(1 - c_{1} \right)\bar{v} + \sqrt{1 - c_{1}^{2}}{\Psi'_{k}},\\
&\alpha_{k}\left[n\right] = c_{2}\alpha_{k}\left[n-1\right] + \left(1 - c_{2} \right)\bar{\alpha} + \sqrt{1 - c_{2}^{2}}{\Psi_{k}},
\end{align}
where $c_1$ and $c_2$ are the influence of MUs’
previous state. $\bar{v}$ and $\bar{\alpha}$ denote the average speed and direction of all users, respectively. $\Psi'_k$ and $\Psi_k$ are two independent Gaussian random variables. 
Based on this, the user's location is updated by $x_{k}\left [ n \right ] =  x_{k}\left [ n-1 \right ] + v_{k}\left [ n-1 \right ]\cos\left ( \alpha_{k}\left [ n-1 \right ] \right )\delta$ and $y_{k}\left [ n \right ] =  y_{k}\left [ n-1 \right ] + v_{k}\left [ n-1 \right ]\sin\left ( \alpha_{k}\left [ n-1 \right ] \right )\delta.$

\subsubsection{UAV Mobility Model}
We design  the position of UAV $u$ in time slot $n$ as ${\bf{q}}_{u}\left [ n \right ] = \left ( x_{u}\left [ n \right ], y_{u}\left [ n \right ] ,z_{u}\left[n\right] \right )$. 
In addition, the velocity and acceleration of the UAV $u$ are defined as
${\bf{v}}_{u}\left[n\right]$ = $\left(v_{u,x}\left[n\right], v_{u,y}\left[n\right], v_{u,z}\left[n\right]\right)$ and ${\bf{a}}_{u}\left[n\right]$, respectively. Consequently, the mobility  of  UAV $u$ is ${\bf{q}}_{u}\left [ n+1 \right ] = {\bf{q}}_{u}\left [ n \right ] + {\bf{v}}_{u}\left [ n \right ]\delta + \frac{1}{2}{\bf{a}}_{}\left [ n \right ]\delta^{2}$.
To enforce feasible operation within the designated area, the UAV positions must satisfy the following boundary constraint:
\begin{align}
    &0 \leq x_{u}\left[n\right] \leq X_{\text{max}}, \ u\in\mathcal{U},n\in\mathcal{N},  \label{x:uav}\\
    &0 \leq y_{u}\left[n\right] \leq Y_{\text{max}}, \ u\in\mathcal{U},n\in\mathcal{N}, \label{y:uav} \\
      &Z_{\text{min}} \leq z_{u}\left[n\right] \leq Z_{\text{max}}, \ u \in \mathcal{U}, n\in\mathcal{N}, \label{z:uav}
\end{align}
where \( X_{\max} \), \( Y_{\max} \) denote the horizontal boundaries of the rectangular area, and \( Z_{\min} \), \( Z_{\max} \) denote the minimum and maximum flight altitudes of the UAVs, respectively.
\subsection{Communication Model}
We consider that the channel between the user and the UAV is modeled as a Rician channel model. Let $g_{k,u} \left[n\right]$ be the channel information between user $k$  and UAV $u$ in time slot $n$, which can be expressed as:
\begin{align}
    g_{k,u} \left[n\right] &= \sqrt{G_{k,u}\left[n\right]} \hat{g}_{k,u} \left[n\right], \label{c:channel}
\end{align}
where $G_{k,u}\left[n\right] = \frac{\beta_{0}} {\left\| {{\bf{q}}_{u}\left[n\right]} - {{\bf{q}}_{k}\left[n\right]} \right\|^2 }$ represents the large-scale average channel power gain and $\beta_{0}$ denotes the average channel power gain.
Moreover, $\hat{g}_{k,u} \left[n\right] = \sqrt{\frac{K_{a}}{K_{a}+1}} \bar{g}_{k,u}\left[n\right] + \sqrt{\frac{1}{K_{a}+1}}\tilde{g}_{k,u}\left[n\right]$  is a complex-valued random variable related to small-scale
fading, where $\bar{g}_{k,u}\left[n\right]$ denotes the deterministic LoS channel component. $\tilde{g}_{k,u}\left[n\right]$ and $K_{a}$ represent the random scattered component and the Rician factor, respectively.

Due to limited bandwidth resources, only a subset of users can be scheduled. 
Let $\beta_{k,u}[n]$ denote the communication association between user $k$ and UAV $u$ in time slot $n$, where $\beta_{k,u}[n] = 1$ indicates that user $k$ is scheduled and associated with UAV $u$. Otherwise, $\beta_{k,u}[n] = 0$.
Considering the orthogonal frequency division multiple access scheme for data transmission, and let $p_{k}$ denote the transmission power of user $k$. 
In time slot $n$, the transmission rate is denoted as:
\begin{align}
R_{k,u} \left[n\right] = \beta_{k,u}\left[n\right]B_{k,u}\left[n\right]\log_2\left( 1+\frac{p_{k}g_{k,u}\left[n\right]}{N_{0}B_{k,u}\left[n\right]}\right),   
\end{align}
where $N_{0}$ represents the noise power spectral density and $B_{k,u}\left[n\right]$ is the bandwidth allocated to user $k$.   

\subsection{Data Sensing and Computing Model}
We denote $\hat{o}_{k}$ as the sensing rate of user $k$, which depends on the type of sensing task.  
Accordingly, the amount of sensing data for user $k$ in time slot $n$ is given as $ D_{k}^{s}\left[n\right] = \xi_{k,1}\left[n\right]\hat{o}_{k}\delta $.
Based on the established communication model, the bit of sensing data offloaded to UAV $u$ is $D_{k,m}^{t}\left[n\right]  = \left(1 - \xi_{k,1}\left[n\right]\right)\xi_{k,2}\left[n\right]R_{k,u}  \left[n\right]\delta$.

Let $f_k[n]$ denote the local CPU frequency of user $k$, and let $C_k$ represent the number of CPU cycles required to process one bit of data.
The amount of data computed locally by user $k$ in time slot $n$ is given as $D_{k,\text{loc}}[n] = \frac{f_{k}[n](1 - \xi_{k,1}[n])\delta}{C_{k}}$.
Similarly, let $f_{k,u}[n]$ denote the computation resources allocated by UAV $u$ to user $k$, and let $C_{u}$ denote the number of CPU cycles per bit at UAV $u$. Then, the amount of data computed by UAV $u$ for user $k$ is $D_{k,u}^{\text{comp}}[n] = \frac{f_{k,u}[n](1 - \xi_{k,1}[n])(1 - \xi_{k,2}[n])\delta}{C_{u}}$.
\subsection{Energy Consumption Model} 
\subsubsection{Energy Consumption at Users}
The total energy consumption of user consists of three components.
First, we denote $e_k$ as the sensing energy per bit, the sensing energy consumption is given by $ E_{k}^{s}[n] = e_k D_k^{s}[n]$. 
Then, the transmission energy consumption can be denoted as $E_{k,u}^{t}\left[n\right] = p_{k}\left(1 - \xi_{k,1}\left[n\right]\right)\xi_{k,2}\left[n\right]\delta$.
Finally, the computing energy consumption is given as $E_{k,\text{loc}}\left[n\right] = \kappa\left(f_{k}\left[n\right]\right)^{3}\left(1 - \xi_{k,1}\left[n\right]\right)\delta$, where $\kappa$ is the effective switched capacitance.
Let $E_{k}\left[n\right]$ denote the total energy consumption, which is obtained as follows:
\begin{align}
    E_{k}\left[n\right] =  E_{k}^{s}\left[n\right] + \sum_{u\in\mathcal{U}} \beta_{k,u}\left[n\right] E_{k,u}^{t}\left[n\right] + E_{k,\text{loc}}\left[n\right]. \label{e:comp}
\end{align}
\subsubsection{Energy Consumption at UAVs}  
The UAV’s energy consumption mainly includes computation-related and flight energy consumption.
The flight power of UAV $u$ is shown in $P_{u}[n]= P_{0}(1 + 3(v_{u,x}[n]^2 + v_{u,y}[n]^2)/(\Omega^2r^2))
+ (P_{i}v_{0}/(v_{u,x}[n]^2 + v_{u,y}[n]^2))
+ \frac{1}{2}d_{0}\rho sA_{r}(v_{u,x}[n]^2 + v_{u,y}[n]^2)^{\frac{3}{2}}+ Gv_{u,z}[n]$ \cite{CaiTWC2022}. 
Therein, the first three components are related to the horizontal flight power. The last component represents the vertical flight power. 
Since the vertical velocity $v_{u,z}\left [ n \right ]$ influences flight endurance and power, it serves as a critical factor in controlling the UAV's flight altitude. 
Consequently, the flight energy consumption of UAV $u$ is denoted as $E_{u}^{p}\left[n\right] = P_{u}\left[n\right]\delta$.
Since user $k$ offloads sensing data to UAV for computation, the corresponding computing energy consumption by UAV $u$ is defined by $E_{k,u}^{\text{comp}}\left[n\right] = \kappa\left(f_{k,u}\left[n\right]\right)^{3}\left(1 - \xi_{k,1}\left[n\right]\right) \left(1 - \xi_{k,2}\left[n\right]\right)\delta$.
Then, we define $E_{u}\left[n\right]$ as the total energy consumption of UAV $u$, which can be expressed as:
\begin{align}
    E_{u}\left[n\right] = \sum_{k\in\mathcal{K}}E_{k,u}^{\text{comp}}\left[n\right] + E_{}^{p}\left[n\right].
\end{align}
Given the limited energy resources of users and UAVs, which are subject to the following constraints: 
\begin{align}
0\leq E_{k} \left [ n \right ] \leq E_{k}^{\max},\ k\in\mathcal{K}, n\in\mathcal{N},\label{p:E_{k}} 
\end{align}
\begin{align}
0 \leq E_{u}\left[n\right] \leq E_{u}^{\text{max}}, \ u\in\mathcal{U}, n\in\mathcal{N},\label{p:Euav} 
\end{align}
where $E_{k}^{\max}$ and $E_{u}^{\max}$ represent the maximum energy consumption of user $k$ and UAV $u$, respectively. 

The objective of our paper aims to maximize the amount of processed sensing data in the multi-UAV-assisted MCS networks. 
To achieve this, we jointly optimize bandwidth allocation ${\bf{B}} \triangleq \{ B_{k,u}[n],k \in \mathcal{K}, u \in \mathcal{U}, n \in \mathcal{N}\}$, the user-UAV associations $\boldsymbol{\beta} \triangleq \{\beta_{k,u}[n], k \in \mathcal{K}, u \in \mathcal{U}, n \in \mathcal{N}\}$, computation resources allocation ${\bf{f}} \triangleq \{f_{k}[n], f_{k,u}[n], k \in \mathcal{K}, u \in \mathcal{U}, n \in \mathcal{N}\}$.
In addition, we consider optimizing UAV acceleration ${\bf{a}} \triangleq \{{\bf{a}}_{u}[n], u \in \mathcal{U}, n \in \mathcal{N}\}$, and the time slot partition $\boldsymbol{\xi} \triangleq \{\xi_{k,1}[n], \xi_{k,2}[n], k \in \mathcal{K}, n \in \mathcal{N}\}$.   
Building upon these definitions, the formulation of the optimization problem is as follows:
\begin{subequations}\label{P0}
    \begin{align}
\max_{{\bf{B}},\boldsymbol{\beta},{\bf{f}},{\bf{a}},\boldsymbol{\xi}} \ &\sum_{n\in\mathcal{N}}\sum_{u\in\mathcal{U}}\sum_{k\in\mathcal{K}}\beta_{k,u}\left [ n \right ]  \left ( D_{k,\text{loc}}\left [ n \right ] + D_{k,u}^{\text{comp}}\left [ n \right ] \right )   \label{problem} \\
            \text{s.t.}\quad
                 &\eqref{x:uav}-\eqref{z:uav}, \eqref{p:E_{k}}-\eqref{p:Euav}, \nonumber\\
                 &\sum_{u\in\mathcal{U}} \beta_{k,u}\left[n\right] \leq 1, k\in\mathcal{K},n\in\mathcal{N}, \label{beta}\\    
                 &\sum_{k\in\mathcal{K}}B_{k,u} \left [n  \right ] \leq B_{u}, u\in\mathcal{U},n\in\mathcal{N}, \label{bandwidth}\\
                 & 0\leq f_{k} \left [n  \right ] \leq f_{k}^{\max}, \ k\in\mathcal{K},n\in\mathcal{N}, \label{p:f_{k}}\\
                 & 0\leq f_{k,u} \left [n  \right ] \leq f_{u}^{\max},\ k\in\mathcal{K},\ u\in\mathcal{U},n\in\mathcal{N}, \label{p:f_{uav}} \\
                 & \sum_{k\in\mathcal{K}} \beta_{k,u}\left[n\right] f_{k,u} \left [n  \right ] \leq f_{u}^{\max}, \ u\in\mathcal{U},n\in\mathcal{N}, \label{p:f_{uav1}}\\
                 & D_{k,\text{loc}} \left [n  \right ] + \sum_{u\in\mathcal{U}} \beta_{k,u}\left[n\right]D_{k,u}^{t}\left [n  \right ] \leq D_{k}^{s}  \left [n  \right ], \label{p:D1} \nonumber\\
                 & k\in\mathcal{K},n\in\mathcal{N}, \\
                 & D_{k,u}^{\text{comp}} \left [ n \right ] \leq D_{k,u}^{t} \left [n  \right ], \ k\in\mathcal{K},u\in\mathcal{U},n\in\mathcal{N}, \label{p:D2} \\ 
                 &\left \| {\bf{v}}_{u}\left [ n \right ] \right \| \leq v_{\max}, \ u\in\mathcal{U},n\in\mathcal{N}, \label{p:v}\\
                 &\left \| {\bf{a}}_{u}\left [ n \right ] \right \| \leq a_{\max}, \ u\in\mathcal{U},n\in\mathcal{N}, \label{p:a}\\
                 & \left \|{\bf{q}}_{u}\left [ n \right ] - {\bf{q}}_{{u}'} \left [ n \right ]\right \| \geq d_{\min}, \nonumber \\
                 &  u\neq {u}', u, u'\in\mathcal{U}, n\in\mathcal{N}, \label{p:collision} \\
    &\xi_{k,1}\left[n\right],\xi_{k,2}\left[n\right] \in \left(0,1\right), \ k\in\mathcal{K},n\in\mathcal{N}, \label{r:r1} 
    \end{align}
\end{subequations}
where 
\eqref{beta} ensures that each user $k$ is associated with at most one UAV.
\eqref{bandwidth} imposes constraints on the available bandwidth resources.
\eqref{p:f_{k}} and \eqref{p:f_{uav}}  limit the allocation of computation resources by users and UAVs, respectively.
\eqref{p:f_{uav1}} enforces the computation capacity constraint at the UAV side. 
\eqref{p:D1} restricts the number of bits processed by user $k$ to be no more than the sensing data.
\eqref{p:D2} indicates the number of data processed by UAV $u$ cannot exceed the data transmitted by user $k$.
\eqref{p:v} and \eqref{p:a} impose limits on the UAV's speed and acceleration, respectively.
\eqref{p:collision} enforces the minimum safety distance to prevent UAV collisions.
\eqref{r:r1} is used to limit the time slot partition variables.
 
The above problem is a mixed-integer nonconvex optimization task, involving both nonconvex objectives and discrete
variables. In addition, it represents a sequential decision-making process under dynamic and uncertain user mobility. Traditional optimization methods are limited by their
reliance on prior knowledge and are not well-suited for such
environments. 
To address these challenges, we propose a novel MADRL algorithm featuring a hybrid actor network architecture.
The details of the proposed approach are presented below.
 
\section{ METHODOLOGY AND PROPERTY ANALYSIS}
In this section, the nonconvex stochastic problem \eqref{P0} is first formulated as a POMDP. To solve it, we then adopt a novel MADRL algorithm named CKAN-HAPPO. 
The proposed method features a redesigned actor network that addresses the limitations of conventional MADRL algorithms, thereby improving overall training performance.

\subsection{POMDP Formulation}
In our proposed model, due to the random mobility and limited sensing capabilities of users, each user can only observe information related to their own tasks and cannot fully observe the global tasks. 
Meanwhile, due to bandwidth limitations, the UAV is unable to receive and process the data from all users simultaneously, and thus it also cannot grasp the complete system state. These factors result in partial observability for both the users and UAVs when making decisions. 
To address the limitation, we model the problem as a POMDP. Formally, a POMDP is defined as 
$\langle \mathcal{I}, \mathcal{S}, \{\mathcal{O}_i\}_{i \in \mathcal{I}}, \{\mathcal{A}_i\}_{i \in \mathcal{I}}, \mathcal{P}, \{\mathcal{R}_i\}_{i \in \mathcal{I}}, \gamma \rangle$, 
where $\mathcal{I} \triangleq \{1, 2, \ldots, K+M\}$ is the set of agents. Then, $\mathcal{S}$ and $\mathcal{O}_i$ represent the state space of all agents and the local state of agent $i$, respectively, i.e., in each step, $\mathcal{S} = \mathcal{O}_1 \times \cdots \times \mathcal{O}_{\mathcal{I}}$. Moreover, $\mathcal{A}_i$ is the action space of agent $i$, $\mathcal{P}$ represents the state transition probability, $\mathcal{R}_i$ is the reward function of agent $i$, and $\gamma \in [0,1]$ denotes the discount factor. The UAV-assisted MCS network consists of two types of agents: users and UAVs. Each agent interacts with the environment in each time step, observing the local state, selecting an action, and receiving a reward. We consider a time step $t$ to be equivalent to a time slot $n$, where both represent discrete time intervals for decision-making and system state transitions.

$\bullet$ \textit{Observation}:
Each agent has access only to partial information of the environment. 
Specifically, the user agent considers not only its own performance, including location, maximum energy consumption, and
sensing rate but also the location of UAVs. The observation of user $k$ is $o_{k}\left [n\right] = \left \{{\bf{q}}_{k}\left [ n \right ], E_{k}^{\text{max}}, \hat{o}_{k}, {\bf{q}}_{u}\left [ n \right ], u\in \mathcal{U} \right \}$.

For UAV agent $u$, the observation  is given by $o_{K+u}[n]=\{{\bf{q}}_{u}\left [ n \right ],{\bf{q}}_{-u}\left [ n \right ], E_{u}^{\text{max}}, {\bf{q}}_{k}\left [ n \right ], k\in\mathcal{K}\}$.
Therein, ${\bf{q}}_{u}\left [ n \right ]$ denotes the UAV postion, ${\bf{q}}_{-u}\left [ n \right ]$ represents the position of other UAVs, $E_{u}^{\text{max}}$ restricts the maximum energy consumption of UAV, and ${\bf{q}}_{k}\left [ n \right ]$ represents the user positions.

$\bullet$ \textit{Action}: After observing the environment, the user determines the time slot partition, computation resource allocation, and communication association in time slot $n$.
The action is defined as $a_{k}\left [n\right]$$ = $$ \left \{ \xi_{k,1}\left[n\right], \xi_{k,2}\left[n\right], f_{k}\left[n\right], \beta_{k,u}\left[n\right], u\in \mathcal{U}   \right \}, $
where $\beta_{k,u}\left[n\right]$ is a binary variable.
To enable tractable optimization, we relax it to between 0 and 1. 

Given limited communication and computation resources, each UAV allocates them to associated users and controls its movement by adjusting acceleration. The action of UAV $u$ is defined as $a_{K+u}\left [n\right] = \left\{{\bf{a}}_{u}\left[n\right], B_{k,u}\left[n\right], f_{k,u}\left[n\right],  k\in\mathcal{K} \right\}$.
\begin{figure}[!t]
\centering
\includegraphics[width=8cm,height=4.6cm]{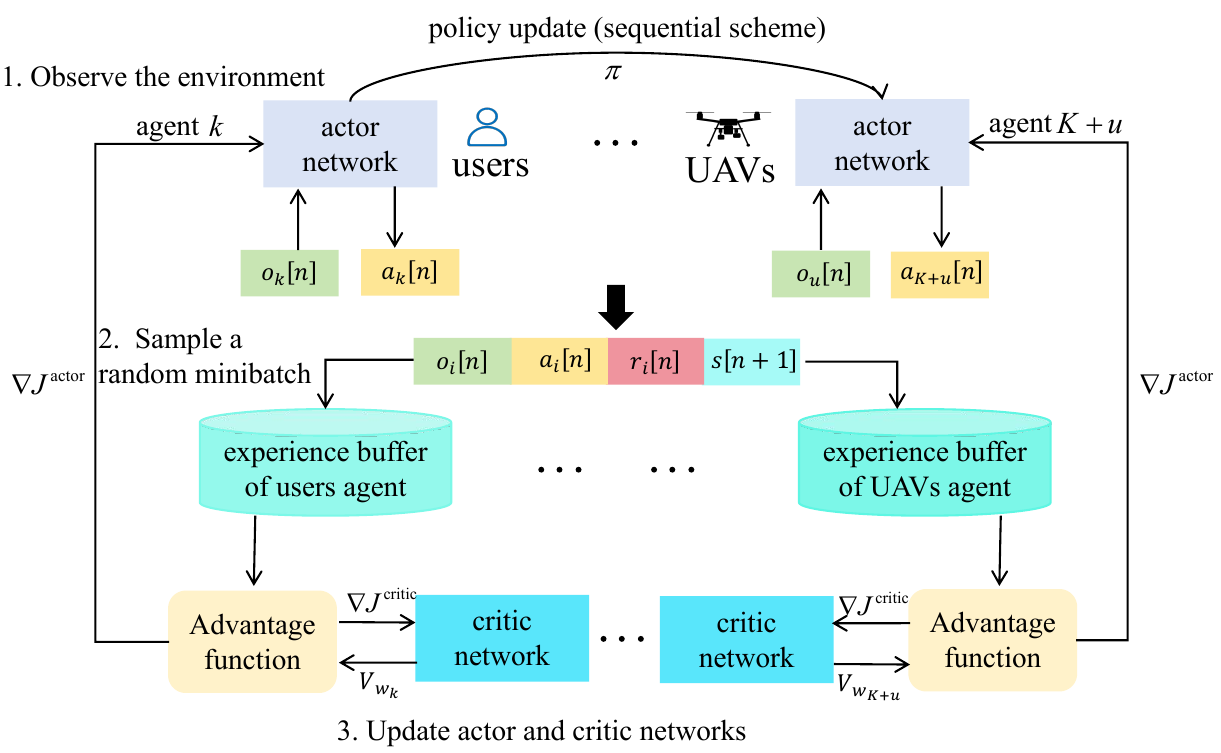}
    \caption{The framework of the CKAN-HAPPO algorithm.} 
\end{figure} 

$\bullet$ \textit{Reward}:
To design an effective policy, the reward function includes
both a goal and penalties for not meeting the constraints. 
The reward function of user $k$ is represented as:
\begin{align}
r_{k}\left [n\right] &= D_{k,\text{loc}}\left[n\right] +\sum_{u\in\mathcal{U}}\beta_{k,u}\left[n\right]D_{k,u}^{\text{comp}}\left[n\right] \nonumber\\
& +P^{\text{e}}_{k}\left[n\right] +P^{\text{c}}_{k}\left[n\right]+P^{t}_{k}\left[n\right].   
\end{align}
For ease of analysis, we define the penalty function as  $F(x,a,b)$$ =$$ -| x - \text{clip}(x,a,b)|$, where $\text{clip}(x,a,b)$ is denoted by  $\text{min}\{ \text{max}\{ x,a \},b\}$. 
Specifically, the penalty for user $k$ exceeding the maximum energy budget is given by $P^{e}_{k}\left[n\right] = \mu^{e}F\left(E_{k}\left[n\right],0,E_{k}^{\text{max}}\right)$. 
Similarly, the penalty for processing more data than the user's sensing capacity is defined as $P^{c}_{k}[n] $$=$$\mu^{c}F(D_{k,\text{loc}}[n] + \sum_{u\in\mathcal{U}}\beta_{k,u}[n]D_{k,u}^{t}[n],0,D_{k}^{s}[n])$.
Furthermore, when the amount of data transmitted by user $k$ is smaller than the  processing capacity of the associated UAV, the penalty incurred by the user is denoted as $P^{t}_{k}[n] $$=$$\mu^{t}F(\sum_{u\in\mathcal{U}}D_{k,u}^{\text{comp}},0,D_{k}^{t}[n])$.
Here, $\mu^{e}$, $\mu^{c}$ and $\mu^{t}$ are the penalty factors.

For UAV agent $u$, the reward function is defined as follows:
\begin{align}
r_{K+u}[n] &= \sum_{k \in \mathcal{K}} \beta_{k,u}[n](D_{k,\text{loc}}[n] + D_{k,u}^{\text{comp}}[n]) \nonumber\\
&+ P_u^e[n] + P_u^t[n] + P_u^r[n] + P_u^c[n],    
\end{align}
where $P_u^e[n]$$ =$$ \overline{\mu}^e F(E_u[n], 0, E_u^{\text{max}})$ penalizes energy overuse, and $P_u^t[n]$$ $$= $$\overline{\mu}^t \sum_{k \in \mathcal{K}} F(D_{k,u}^{\text{comp}}[n], 0, D_{k,u}^t[n])$ reflects mismatches between computed and transmitted data. The term $P_u^r[n]$$ =$$ \overline{\mu}^r F(\mathbf{q}_u^h[n], 0, Z)$ imposes a penalty for operating beyond the designated area, where $\mathbf{q}_u^h[n]$$ =$$ (x_u[n], y_u[n])$ denotes the UAV’s horizontal position and $Z$ $= $$(X_{\text{max}}, Y_{\text{max}})$ defines the boundary. In addition, $P_u^c[n]$$ =$$ \overline{\mu}^c \sum_{u' \ne u} F(\|\mathbf{q}_u[n] - \mathbf{q}_{u'}[n]\|, 0, d_{\text{min}})$ enforces a minimum separation distance between UAVs. All penalty terms are weighted by the factors $\overline{\mu}^e$, $\overline{\mu}^t$, $\overline{\mu}^r$, and $\overline{\mu}^c$.

To simplify model training, we further reduce the action space dimension. Specifically, the computation frequency of user $k$ in slot $n$ can be obtained analytically, as shown in the following theorem.
\begin{theorem}
    Based on the obtained parameters ${\bf{B}}$, $\boldsymbol{\beta}$, ${\bf{a}}$, $\boldsymbol{\xi}$, and $f_{k,u}[n]$, the optimal computation frequency for user $k$ in time slot $n$ is represented as $f_k^*[n]$. The value can be directly derived as $f_{k}^{*}[n] = \max(0,\min(f_{k}^{\max}, f_{k}^{1}, f_{k}^{2}))$, where $f_{k}^{1} = \frac{\left(D_{k}^{s}[n] - \sum_{u\in\mathcal{U}}\beta_{k,m}[n]D_{k,u}^{\text{t}}[n]\right)C_{k}}{(1 -\xi_{k,1}[n])\delta}$, and $f_{k}^{2} = \sqrt[3]{\frac{(E_{k}^{\text{max}} -E_{k}^{s}[n]-\sum_{u\in\mathcal{U}}\beta_{k,u}[n]E_{k,u}^{\text{t}}[n])}{k\left(1-\xi_{k,1}[n]\right)\delta}}$. \\
\end{theorem}
\begin{IEEEproof}
The proof is omitted due to space limitations.
\end{IEEEproof}  

\begin{algorithm}[t]
    \caption{ Proposed CKAN-HAPPO training framework}
    \begin{algorithmic}[1] 
    \STATE Initialize the maximum number of 
     episodes Max\_e, episode length Epi\_l, and the number of updates during training, i.e., p\_e.
    \STATE For each agent $i \in \mathcal{I}$, initialize the actor network $\theta_i$, critic network $w_i$, and the experience buffer $B_i$.
    \FOR{episode = 1,$\dots$, Max\_e} 
        \FOR {$n$ = 1,$\dots$, Epi\_l} 
        \STATE Obtain observations $o_{i}\left[n\right]$ from the environement and get actions $a_{i}\left[n\right],i\in\mathcal{I}$.
        \STATE Calculate the value $f_{k}^{*}\left[n\right]$ based on Theorem 1.
        \STATE Calculate the reward $r_{i}\left[n\right],i\in\mathcal{I}$, move to the next state and store the transition into buffer $\mathcal{B}_{i}$. 
        \ENDFOR
        \STATE Sample a random minibatch from $\mathcal{B}_{i},i\in\mathcal{I}$.
        \STATE Shuffle agents randomly to form a new set, i.e., $\mathcal{I^{'}}$.
        \FOR{epoch = 1,$\dots$, p\_e}
        \FOR{agent ${i'}$ in $\mathcal{I}^{'}$}
        \STATE Update actor network $\theta_{{i'}}$ base on \eqref{lossactor} and compute $M_{\boldsymbol{\pi}}^{{1:i'+1}}\left(s\left[n\right],\hat{{\bf{a}}}\left[n\right]\right)$. 
        \ENDFOR
        \STATE Update critic network $w_{{i'}}$ based on \eqref{critic loss}.
        \ENDFOR
    \ENDFOR   
    \end{algorithmic}
\end{algorithm}

\subsection{MADRL Algorithm}
Building on the POMDP formulation, we propose a novel MADRL algorithm named CKAN-HAPPO to solve the problem.
This algorithm is built upon the CTDE paradigm, which effectively mitigates the non-stationarity inherent in multi-agent environments. 
\subsubsection{Neural Network Design} 
Before training and execution, we construct a novel hybrid architecture consisting of a one-dimensional CNN module and a KAN module. 
The CNN module is utilized to extract representative features from agent observations due to  its feature extraction capability, while the inherent parameter-sharing property of convolutional filters contributes to reduced model complexity \cite{wu2020optimal}. 
Following the convolutional layers, the KAN module serves as the action generation component, replacing conventional MLP networks. By employing learnable spline-based activation functions, the KAN module enables more accurate modeling of complex nonlinear relationships between states and actions with fewer trainable parameters \cite{liu2024kan}. 
This hybrid design enhances both the learning performance and the generalization ability in complex environments. 
\subsubsection{CKAN-HAPPO Algorithm} 
CKAN-HAPPO adopts a sequential update scheme,
as illustrated in Fig. 2.
Each agent maintains its own actor and critic networks, parameterized by $\theta$ and $w$, respectively. 
The actor generates actions, while the critic evaluates the value function. 
A shared policy $\boldsymbol{\pi}$ is jointly optimized across agents.
During training, agents first observe the environment to get actions.
Subsequently, their local observations $o_{{i}}[n]$ are aggregated into a global state $s[n]$, which, together with actions $a_i[n]$, is used to compute rewards $r_i[n]$. 
The transitions $\{o_{i}[n],a_i[n],r_i[n],s[n+1] \}$ is stored in the experience buffer.
A minibatch is then sampled to update the networks using gradient-based loss functions $J(\theta)$ and $J(w)$.   After each update, the actor parameters are synchronized back to the agents.

We incorporate the principle of multi-agent advantage function decomposition into the actor network to improve global coordination through coordinated policy updates  \cite{Kuba2021TrustRP}.  
Specifically, let 
$\hat{{\bf{a}}}\left[n\right]$ and $\theta_{{i}}$
 denote the agents’ joint action strategy in time slot 
 $n$ and the parameters of agent $i$ actor network, respectively.
 The actor-network loss function is then given by: 
\begin{align}
    &J\left(\theta_{{i}} \right) = 
    \mathbb{E}\Bigg[  \min\Bigg( 
    \frac{\pi_{\theta^{{i}}_{\text{new}}}\left( a_{{i}}\left[n\right] \middle| o_{{i}}\left[n\right] \right)}
    {\pi_{\theta^{{i}}_{\text{old}}}\left( a_{{i}}\left[n\right] \middle| o_{{i}}\left[n\right] \right)} 
    M_{\boldsymbol{\pi}}^{{1:i}}\left( s\left[n\right], \hat{{\bf{a}}}\left[n\right] \right), \notag \\
    &\text{clip}\Bigg( 
    \frac{\pi_{\theta^{{i}}_{\text{new}}}\left( a_{{i}}\left[n\right] \middle| o_{{i}}\left[n\right] \right)}
    {\pi_{\theta^{{i}}_{\text{old}}}\left( a_{{i}}\left[n\right] \middle| o_{{i}}\left[n\right] \right)}, 
    1 \pm \epsilon \Bigg) 
    M_{\boldsymbol{\pi}}^{{1:i}}\left( s\left[n\right], \hat{{\bf{a}}}\left[n\right] \right) 
    \Bigg) \notag \\
    & + \psi S\left( o_{{i}}\left[n\right] \right) 
    \Bigg], \label{lossactor}
\end{align}
where $\pi_{\theta^{i}_{\text{old}}}$ and $\pi_{\theta^{i}_{\text{new}}}$ denote the policy of agent $i$ before and after the update, respectively, and $\psi S(o^i[n])$ represents the policy entropy that encourages exploration. 
Moreover, the clip operator restricts the old-to-new strategy ratio between $1-\epsilon$ and $1+\epsilon$  to prevent large policy updates.
Therein, $M_{\boldsymbol{\pi}}^{{1:i}}(s[n], \hat{{\bf{a}}}[n]) 
= \frac{\pi_{\theta_{\text{new}}^{{1:i-1}}} 
\left(\hat{{\bf{a}}}_{{1:i-1}}[n] \middle| o_{{1:i-1}}[n] \right)}
{\pi_{\theta_{\text{old}}^{{1:i-1}}} 
\left(\hat{{\bf{a}}}_{{1:i-1}}[n] \middle| o_{{1:i-1}}[n] \right)} \hat{A}^{{i}}(s[n], \hat{{\bf{a}}}[n])$ reflects the sequential update program of the algorithm,
where $\hat{A}^{{i}}\left(s[n],\hat{{\bf{a}}}[n]\right)$ is an estimate of the advantage function.
The generalized advantage estimation (GAE) is used to equalize the value between variance and bias, then $\hat{A}^{{i}}\left(s\left[n\right],\hat{{\bf{a}}}\left[n\right]\right) 
=  \sum_{l=0}^{\infty} \left(\gamma\lambda\right)^{l} \big(r_{{i}}\left[n + l\right] + \gamma V_{{i}}\left(s\left(n + l + 1\right)\right)   
+ V_{{i}}\left(s\left[n + l\right]\right) \big)$.
Therein, $\lambda\in\left(0,1\right)$ is the generalized advantage estimator and  $V_{{i}}\left(s\left[n\right]\right) = \sum_{l=0}^{\infty}\gamma^{l}r_{{i}}\left[n+l\right]$ is the cumulative discounted
reward, which also represents the state-value function.

To introduce the critic's loss function, we define $V_{w_{{i}}}\left(s\left[n\right]\right)$ to denote the state value function estimated by the critic network. Consequently, the critic network can be updated using the following loss function \cite{10680286}:
\begin{align}
    J\left(w_{{i}}\right) = \frac{1}{2}\left(V_{{i}}\left(s\left[n\right]\right) - V_{w_{{i}}}\left(s\left[n\right]\right) \right)^{2}. \label{critic loss}
\end{align}
Based on the analysis of the above loss functions, we update the actor and critic networks using gradient ascent and gradient descent methods according to \eqref{lossactor} and \eqref{critic loss}, respectively, i.e:
\begin{align}
    &\theta_{{i}} = \theta_{{i}} + l_{a} \nabla \left( J\left( \theta_{{i}} \right) \right), \\
    &w_{{i}} = w_{{i}} - l_{c} \nabla \left( J\left( w_{{i}} \right) \right),
\end{align}
where $l_{a}$ and $l_{c}$ denote the learning rate of the actor and critic network, respectively.

We summarize the training framework of the CKAN-HAPPO algorithm in Algorithm 1. Lines 3-8 describe the interaction between the agents and the environment, where the results  are stored in the experience buffer for future use. 
Lines 9–10 involve sampling mini-batches from the buffer to facilitate network parameter updates.
Lines 11–17 describe the update procedure for both the actor and critic networks.
By iteratively executing this process over multiple episodes, the agent policies are progressively refined and redeployed for continuous interaction with the environment.
\begin{figure}[t!]
\begin{minipage}[t]{0.45\linewidth} \centerline{\includegraphics[scale=0.185]{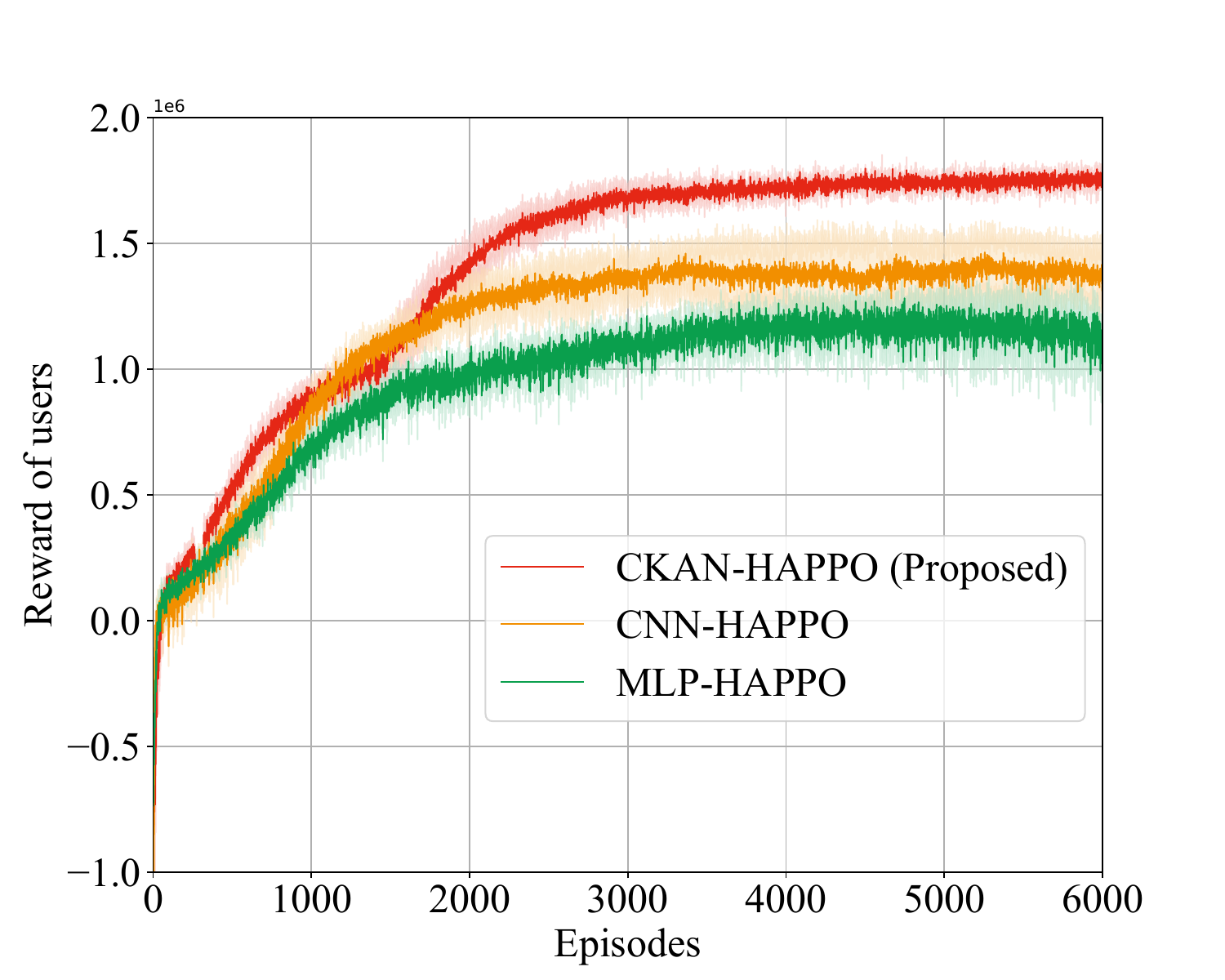}}
    \caption{The convergence of rewards for user agent.}
    \label{reward_users}
\end{minipage}%
    \hfill%
\begin{minipage}[t]{0.45\linewidth} \centerline{\includegraphics[scale=0.185]{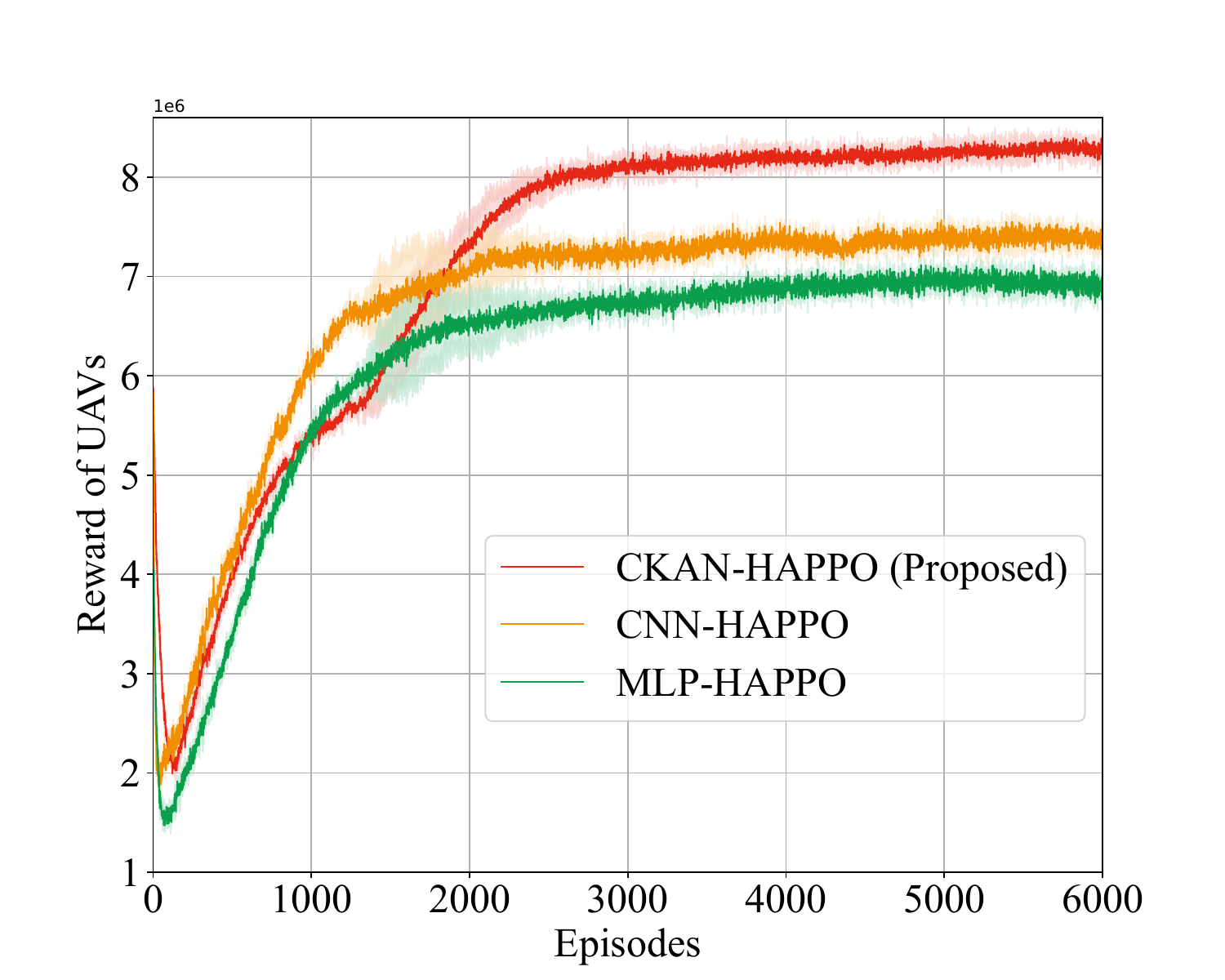}}
    \caption{The convergence of rewards for UAV agent.}
    \label{reward_UAVs}
\end{minipage} 
\end{figure}
\subsection{Complexity Analysis}  
Since actor and critic networks have different
structures, we analyze their complexity respectively.
The actor network is composed of CNN and KAN.
The computational complexity of the $J_{a}$ layer convolutional layer in the CNN is  $\mathcal{O}(\sum_{j=2}^{J_{a}-1}\hat{M}_{j}^{2}\hat{K}_{j}^{2}C_{j-1}C_{j})$, where $\hat{M}_{j}$ and $\hat{K}_{j}$ denote the dimension of the output and the size of the convolution kernel in $j$-layer, respectively. 
$C_{j-1}$ and $C_{j}$ denote the number of input and output channels of the $j$-th layer, respectively.  
KAN serves as a linear layer in the actor network.
The computational complexity of its $J_{a'}$ layer is $\mathcal{O}(\sum_{j=2}^{J_{a'}-1} G'_{j-1}\chi_{j-1}\chi_{j} + G'_{j}\chi_{j}\chi_{j+1})$, where $\chi_{j}$ denotes the number of neurons in the $j$-th layer, and $G'_{j}$ represents the number of spline segments used for piecewise approximation.
Furthermore, $f_{k}\left[n\right]$ can be obtained directly with complexity $\mathcal{O}\left(K\right)$. The computational complexity of the actor network is defined as $\varrho_{a}$, denoted as $\varrho_{a} = \mathcal{O}(\sum_{j=2}^{J_{a}-1} \hat{M}_{j}^{2} \hat{K}_{j}^{2} C_{j-1} C_{j} + \sum_{j=2}^{J_{a'}-1} G'_{j-1} \chi_{j-1} \chi_{j} + G'_{j} \chi_{j} \chi_{j+1} + K 
)$. 
The critic network uses MLP, and the computational complexity of the $J_{c}$ layer is $\mathcal{O}(\sum_{j=2}^{J_{c}-1} \chi_{j-1}\chi_{j} + \chi_{j}\chi_{j+1})$ \cite{10680286}.  The computational complexity of Algorithm 1 is $\mathcal{O}\left(\text{Max\_e} \cdot \text{Epi\_l} \left( \varrho_{a} + \varrho_{c} \right)   \right)$.

\begin{figure}[t!]
\begin{minipage}[t]{0.45\linewidth} \centerline{\includegraphics[scale=0.185]{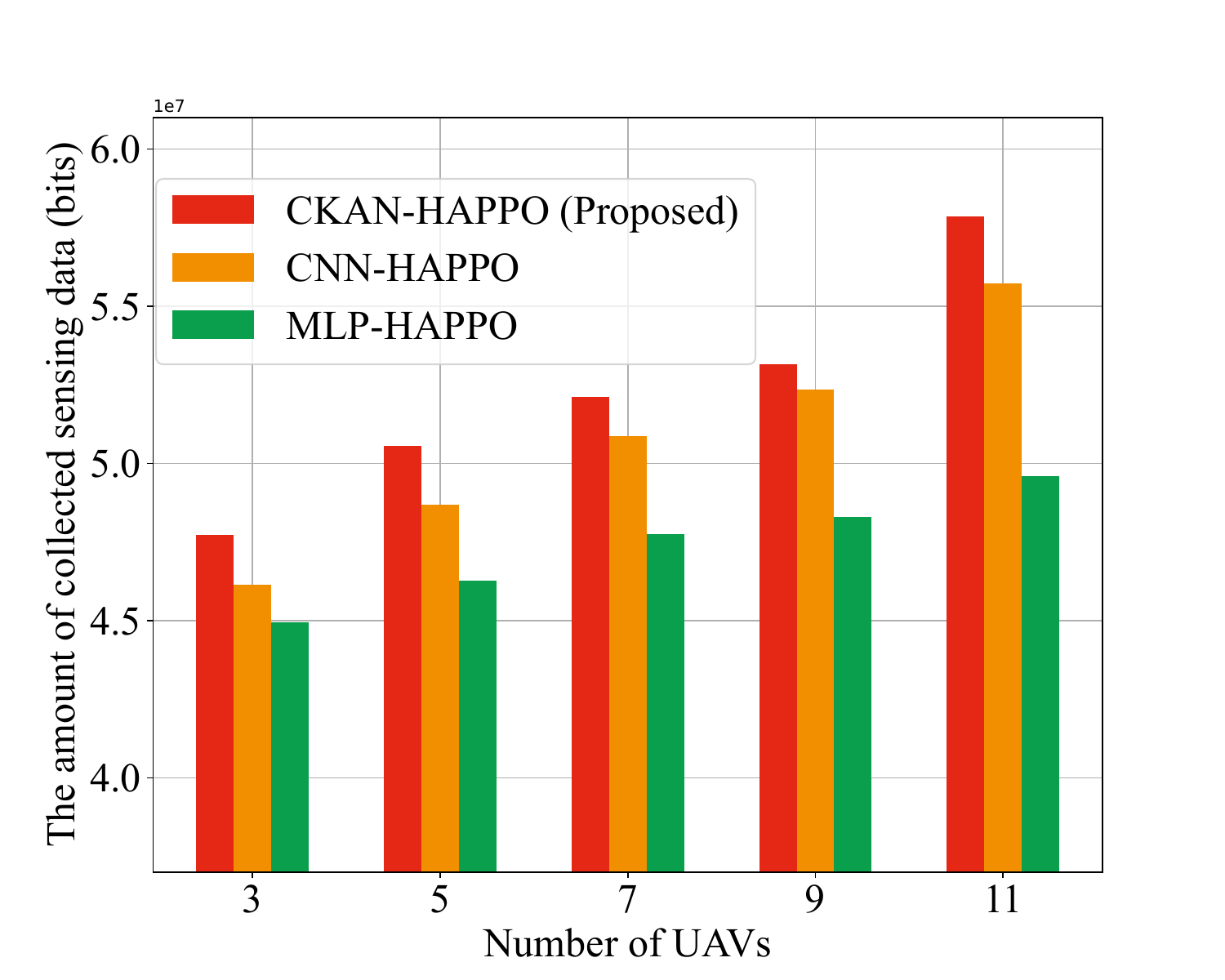}}
    \caption{{The impact of the number of UAVs.}} 
    \label{number_uavs}
\end{minipage}%
    \hfill%
\begin{minipage}[t]{0.48\linewidth} \centerline{\includegraphics[scale=0.185]{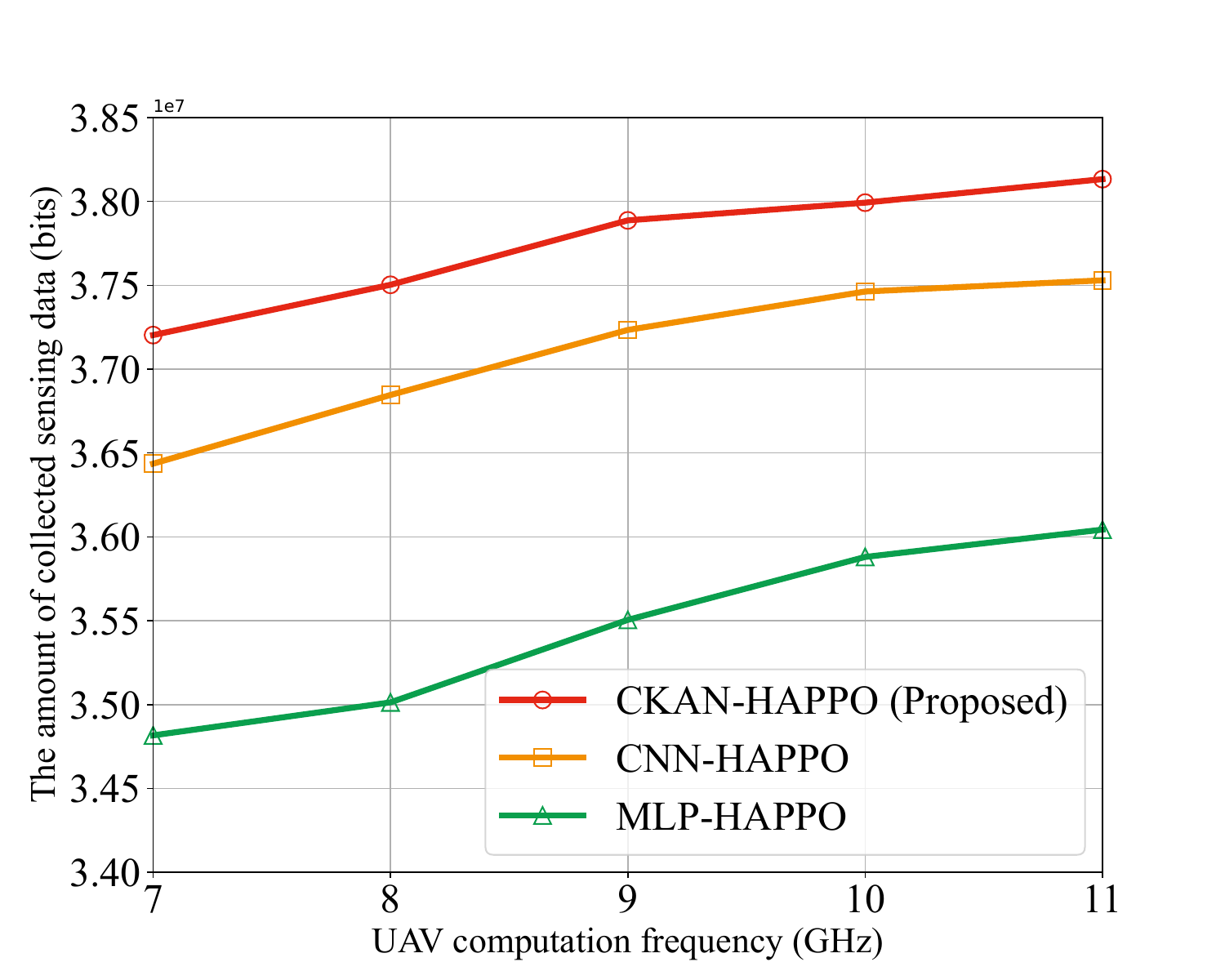}}
    \caption{ The impact of the UAV computation frequency.} 
    \label{computation_frequency}
\end{minipage} 
\end{figure}  
\section{NUMERICAL SIMULATIONS}
This section presents the performance evaluation of the proposed CKAN-HAPPO algorithm. 
We consider a 1000 $\times$ \SI{1000}{m^{2}} square area with $K = 20$ users and $U = 5$ UAVs. 
In the system, the time $T$ = \SI{20}{s} is divided into $N$ time slots with $\delta$ = \SI{1}{s}.   
The sensing data rate \(\hat{o}_{k} \in [5, 10]\,\text{Mb/s}\) and the number of CPU cycles required per bit \(c_k,c_u \in [100, 500]\,\text{cycles/bit}\).
The maximum computation frequency of user and UAV is \SI{1}{GHz} and \SI{8}{GHz}, respectively.
The total bandwidth for UAV $u$ is set to $B_u = $ \SI{10}{MHz}.
UAVs operate at altitudes constrained to \SIrange{100}{300}{m}, with a maximum flight speed and acceleration of $v_{\max} = \SI{30}{m/s}$ and $a_{\max} = \SI{3}{m/s^{2}}$, respectively.
The UAV settings $P_0$, $P_i$, $s$ and $v_{0}$ is set as \SI{79.86}{W}, \SI{88.63}{W}, \SI{0.05}{m^{2}} and \SI{3.6}{m/s}, respectively.
Simulation configurations are aligned with the frameworks proposed in \cite{9583941}\cite{10571223}. 
We compare the performance of the proposed scheme with the benchmarks as follows:

$\bullet$ CNN-HAPPO: In the scheme, the actor network uses the same convolutional architecture as our method but excludes the KAN module, serving as an ablation method \cite{10713326} .

$\bullet$ MLP-HAPPO: Under the scheme, the actor network uses the MLP structure, which is also known as the original HAPPO algorithm \cite{Kuba2021TrustRP}.

Fig. \ref{reward_users} and Fig. \ref{reward_UAVs} show the convergence of user and UAV rewards, respectively. In both cases, the rewards gradually increase with training episodes and eventually stabilize, indicating that the proposed MADRL framework achieves good convergence. 
Among all methods, CKAN-HAPPO achieves the highest rewards, benefiting from its CNN-KAN-based actor network. Compared to MLP, CNN offers stronger feature extraction, while KAN models complex nonlinear state-action mappings more effectively through learnable spline functions. This joint architecture enhances exploration and improves training performance in high-dimensional continuous action spaces.

Fig. \ref{number_uavs} shows the impact of the number of UAVs on the amount of processed sensing data, with $20$ users. 
The results illustrates that the total amount of processed sensing data increases with the number of UAVs. This improvement is attributed to the enhanced computational capabilities brought by UAVs, which can process more sensing data from users. However, the performance gain diminishes as the number of UAVs continues to grow. This is primarily due to the fixed communication bandwidth between users and UAVs, which constrains the volume of sensing data that can be transmitted. Despite this limitation, the proposed algorithm consistently achieves the highest data processing performance, demonstrating its efficiency.

Fig. \ref{computation_frequency} illustrates the relationship between UAV computation frequency and the amount of processed sensing data, with the number of users, UAVs, and bandwidth are fixed at $5$, $20$, and
\SI{10}{MHz}, respectively. 
As the computation frequency increases, more sensing data can be processed, and CKAN-HAPPO consistently outperforms other methods. However, the improvement becomes less pronounced at higher frequencies due to fixed sensing rates and limited bandwidth, which constrain the amount of data that can be transmitted. This leads to a saturation effect in the data processed by UAVs.

\section{Conclusion}
In this paper, we investigated a multi-UAV-assisted MCS network aiming to maximize the amount of processed sensing data. 
The problem was modeled as a joint optimization of time slot partition, user-UAV association, resource allocation, and UAV trajectory. Due to the randomness in the network, the problem became a non-convex sequential decision-making task in a dynamic environment.
To address this, we reformulated the problem as a POMDP and proposed a novel MADRL algorithm called CKAN-HAPPO.
The proposed method integrated CNN and KAN into the actor network to enhance training performance with fewer parameters. 
Furthermore, we reduced the input dimensionality of the actor network by deriving closed-form expressions for user computation resources.
Simulation results demonstrated that our proposed algorithm significantly outperforms existing baselines. 
\bibliographystyle{IEEEtran}
\bibliography{ref}

\end{document}